\documentclass[conference]{IEEEtran}
\IEEEoverridecommandlockouts
\usepackage{cite}
\usepackage{amsmath,amssymb,amsfonts}
\usepackage{algorithmic}
\usepackage{graphicx}
\usepackage{array}
\usepackage{textcomp}
\usepackage{multirow}
\usepackage{xcolor}
\def\BibTeX{{\rm B\kern-.05em{\sc i\kern-.025em b}\kern-.08em
    T\kern-.1667em\lower.7ex\hbox{E}\kern-.125emX}}

\begin{document}

\title{Hybrid AI-based Anomaly Detection Model using Phasor Measurement Unit Data}

\author{\IEEEauthorblockN{Yuval Abraham Regev\IEEEauthorrefmark{1},
Henrik Vassdal\IEEEauthorrefmark{1},
Ugur Halden\IEEEauthorrefmark{1}, 
Ferhat Ozgur Catak\IEEEauthorrefmark{2},
Umit Cali\IEEEauthorrefmark{1}}
\IEEEauthorblockA{\IEEEauthorrefmark{1}Norwegian University of Science and Technology, Høgskoleringen 1, 7491 Trondheim, Norway
}
\IEEEauthorblockA{
\IEEEauthorrefmark{2}University of Stavanger, Kjell Arholms gate 41, 4021 Stavanger, Norway
}
}

\maketitle

\begin{abstract}

Over the last few decades, extensive use of information and communication technologies has been the main driver of the digitalization of power systems. Proper and secure monitoring of the critical grid infrastructure became an integral part of the modern power system. Using phasor measurement units (PMUs) to surveil the power system is one of the technologies that have a promising future. Increased frequency of measurements and smarter methods for data handling can improve the ability to reliably operate power grids. The increased cyber-physical interaction offers both benefits and drawbacks, where one of the drawbacks comes in the form of anomalies in the measurement data. The anomalies can be caused by both physical faults on the power grid, as well as disturbances, errors, and cyber attacks in the cyber layer. This paper aims to develop a hybrid AI-based model that is based on various methods such as Long Short Term Memory (LSTM), Convolutional Neural Network (CNN) and other relevant hybrid algorithms for anomaly detection in phasor measurement unit data. The dataset used within this research was acquired by the University of Texas, which consists of real data from grid measurements. In addition to the real data, false data that has been injected to produce anomalies has been analyzed. The impacts and mitigating methods to prevent such kind of anomalies are discussed. 
\end{abstract}


\section{Introduction REWRITTEN}
With the ever increasing electricity demand and higher share of intermittent renewables in the energy mix due to climate targets \cite{IntergovernmentalPanelonClimateChange2015DriversMitigation}, operating and maintaining a functional electrical grid is becoming increasingly complex \cite{Villamor2020}. Thus, to ensure a reliable power grid, increased monitoring and data handling capacity is needed.

The introduction of PMUs have increased the potential for more frequent measurements and in return, the potential for grid state estimation \cite{ZANNI2020106649}. Traditional Supervisory Control and Data Acquisition (SCADA) systems collect and monitor various data from Remote Terminal Units (RTUs), which are done as scans over a time interval. These RTUs collect data regarding the voltage magnitude and angle, as well as the active and reactive powers at their locations, which provides a good overview and indication about the power grid. However, since RTUs are not time synchronized, in the case of an anomalous event, the measurements might not be relevant \cite{9087782}. Therefore, utilizing time syncronized PMUs has great potential \cite{Karpilow2020DetectionTechniques}.

PMUs are able to provide extreme near real time measurements within the power grid by utilizing a measuring frequency coupled with the grid frequency or other frequencies which are extremely close, which allows for good anomaly detection and continuous state estimations with dynamic observability \cite{MironovEffectSystem}. However, millions of datapoints can be accumulated quickly due to high measurement frequency of multiple grid parameters. PMU datasets have been shown to be vulnerable to cyber attacks such as False Data Injections (FDIs) and adversarial machine learning attacks, either in the form of intentional attacks such as stealth attacks which aims to deliver normal measurements while the supply is being attacked \cite{Khalafi2022IntrusionNetworks}, fast-gradient sign method attack which aims to maximize the loss function or unintentional errors in the measurements \cite{Khare2021AMeasurements,Lee2014CyberAlgorithm}. It should be noted that it is especially the PMU's networks and protocols which are prone to FDIs and leads to decision makers relying on false data \cite{Khalafi2022IntrusionNetworks}.


There are many types of faults and anomalies that occur on the power system, and each anomaly has a consequence of varying severity. In the occurrence of a fault, one of the biggest threats to the system operator is the failure to adjust according to the instantaneous electrical demand. Since, electricity is an essential commodity in our society, failing to supply electricity can have grave political, social and economic consequences. The failure to supply is experienced as anything from load shedding of small areas and/or industry, to large scale blackouts. 
As businesses increasingly become more digital, the economy becomes more sensitive to anomalies in the power system. Additionally, the economy doesn't only react to large scale blackouts, but also when there are issues with the Power Quality (PQ). Such issues include everything that can result in deviations from nominal power frequency and voltage levels such as voltage and current surges, sags, harmonics and transients \cite{Lineweber2001TheCompanies}. In the U.S. alone the economy is reportedly loosing 104-164 billion USD a year due to power outages and major blackouts. In addition, another 15-25 billion USD is lost to lower PQ each year \cite{Lineweber2001TheCompanies}.

In \cite{Walker2014CountingTechnology}, the effects of electricity outfalls in the UK was studied. For the study, the authors conducted a survey from which the authors could report that people, in general can cope with short disruptions. The social impacts will be minimal as long as the outages lasts less then 24 hours in total. However, in the future, the impacts of blackouts might increase due to our increasing reliance on electric and rapid digitalization of the services. Electricity demand is known for being extremely inelastic \cite{Walker2014CountingTechnology}. Hence, electricity shortfall will result in rising power prises, which will in return increase fuel poverty and can put major pressure on any sitting governments across the different nations.  

The main contributions of this paper can be summarized as follows: 1) Comprehensive review of various types of ML/AI models that are used for identification of anomalies in the power systems domain, 2) Development of a hybrid AI-based anomaly detection model using PMU data, and 3) Comparative evaluation of the investigated options and their results as a sensitivity analysis. 

\section{Background}
The power system is a vast framework containing components such as loads, generators, transformer stations, transmission lines and other electrical installations whose job is to produce and transport electricity in an economically efficient way \cite{Cali2021DigitalizationInformatics}. Utilizing PMUs as a part of the data acquisition process have the potential to provide a more accurate description of
what state the power system is in \cite{SrikumarM.S.2015LineUnits}.

\subsection{PMUs and Power Systems}
Transmission grid is the most essential part within a power system as it transfers vast amounts of electricity between the generator and load hubs within countries and across the borders.

PMUs are devices that effectively measure the positive sequence voltages and currents. Due to its precision, the voltage phase angle and rate of change in frequency can be calculated. In addition, utilizing Global Positioning System (GPS) communication to time synchronize the measurements (synchrophasor) via a phase locked sampling signal across a large part of the grid allows for faults to be detected  \cite{ArunG.PHADKE2018PhasorSystems}. 

The measurements from PMUs come in the form of phasors. A phasor is a vector representation of a voltage or current. A sinusoidal wave is given by:

\begin{equation}
    u(t)=U_{m}cos(\omega t + \delta_{0})
\end{equation}

The corresponding representation of the synchrophasor is given by:

\begin{equation}
    U:= \frac{U_{m}}{\sqrt{2}} e^{j\delta_{0}}=\frac{U_{m}}{\sqrt{2}}(cos(\delta_{0}) + jsin(\delta_{0})) = Re\{U\} + jIm\{U\}
\end{equation}

Where the factor $\frac{1}{\sqrt{2}}$ represents the root mean square value (RMS) of the sinusoid, $\omega=2\pi f$ is the angular frequency, where f is the frequency and j is the imaginary number. $\delta_{0}$ is the phase angle of the signal with respect to a given reference point, while $U_{m}$ is the magnitude of the signal.

The PMUs ability to synchronize each measurement over a large area using GPS deems it a promising technology for the future power systems \cite{Vicol2013MODERNMONITORING}. The term synchrophasor describes the phasors that have been estimated at a given time stamp, whereas the time stamps are created by the PMU at multiples of the power systems' nominal frequency \cite{Ree2010SynchronizedSystems}.

\subsection{CNN}

Convolutional Neural Networks (CNN) are specifically designed to manage variations in two dimensional shapes \cite{LeCun1998Gradient-basedRecognition}. However, the techniques' ability to bring out spatial characteristics in a data set makes it an interesting candidate for anomaly detection in one dimensional time series data as well \cite{Chung2022DistrictAttention}.

\begin{figure}[t]
    \centering
    \includegraphics[width=\linewidth]{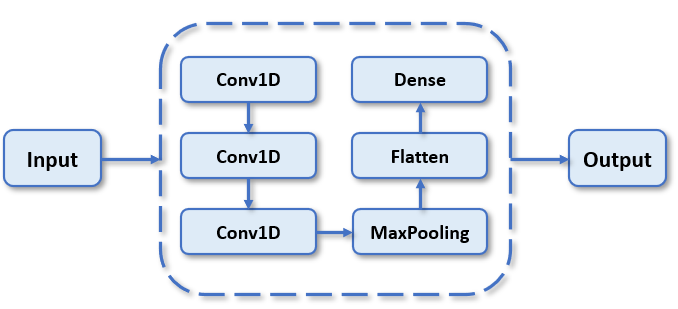}
    \caption{Proposed architecture of the CNN model.}
    \label{fig:CNN flowchart}
\end{figure}

The proposed architecture of the CNN model is illustrated in Figure \ref{fig:CNN flowchart}. The model contains three CNN layers which are followed by a max-pooling layer to save the most important signals. Furthermore, the signals are flattened to a single dimension before a fully connected dense layer is implemented to interpret the extracted features, as similarly performed by \cite{Agga2022CNN-LSTM:Production}.

\subsection{LSTM}
Long Short-Term Memory (LSTM) is a type of deep neural network algorithms that is popular and well-suited for time series data analysis \cite{Lindemann2021ANetworks}. The architecture consists of an input layer, one or more hidden layers, memory cells and an output layer. Using memory and forget cells, the algorithm is capable of preserving long term temporal relationships and patterns in the data.

\begin{figure}[b]
    \centering
    \includegraphics[width=\linewidth]{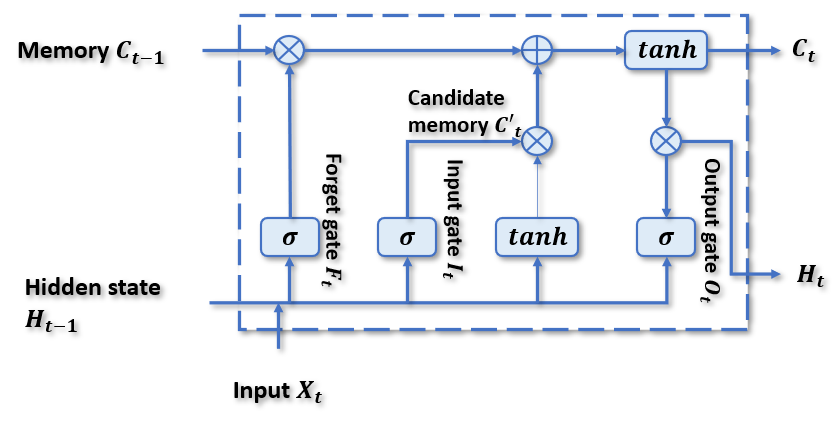}
    \caption{LSTM data flow.}
    \label{fig:LSTMdataflow}
\end{figure}

Figure \ref{fig:LSTMdataflow} shows the data flow in the LSTM model with three different gates utilized in LSTM as; The forget gate, $\mathbf{F_t}$, input gate, $\mathbf{I_t}$ and output gate, $\mathbf{O_t}$ which are designed to better capture long range dependencies in the sequences. The inputs and outputs of each gate has been modelled as at the time step "t", hence the equations for each gate can be written as:

\begin{equation}
    \mathbf{I_t} = \sigma\left(\mathbf{X}_t\mathbf{W}_{xi} + \textbf{H}_{t-1}\mathbf{W}_{hi}+\mathbf{b}_i\right)
    \label{Eq:It}
\end{equation}
\begin{equation}
    \mathbf{F_t} = \sigma\left(\mathbf{X}_t\mathbf{W}_{xf} + \textbf{H}_{t-1}\mathbf{W}_{hf}+\mathbf{b}_f\right)
    \label{Eq:Ft}
\end{equation}
\begin{equation}
    \mathbf{O_t} = \sigma\left(\mathbf{X}_t\mathbf{W}_{xo} + \textbf{H}_{t-1}\mathbf{W}_{ho}+\mathbf{b}_o\right)
    \label{Eq:Ot}
\end{equation}

Equations \ref{Eq:It}, \ref{Eq:Ft} and \ref{Eq:Ot} show how the output value of each gate is computed. Here $\sigma$ represents the activation function, $\mathbf{W}$, the weight parameters  and  $\mathbf{b}$, the bias parameters. $\mathbf{X_t}$ and $\mathbf{H}_{t-1}$ are the input value to the layer and the output of the hidden layer of the previous step, respectively.

\subsection{C-LSTM}
Convolutional Long Short Term Memory (C-LSTM) is one of the hybrid ML algorithms which was utilized and tested in this paper. The main idea behind this hybridization is to combine two different architectures to achieve a better result, since utilization of both algorithms in a combined way can capture both the temporal and spatial features of the data. 

\subsection{Bi-LSTM}

Bidirectional LSTM (Bi-LSTM) has the same properties and structure of normal LSTM layers. However, a Bi-LSTM layer equals two normal LSTM layers where one of the layers operate as normal in the forward direction, whereas the other layer operates in the opposite and backwards direction. This type of connection results in a model which is better equipped for understanding the context of the data points, as it looks to the data from both directions. 

\subsection{Evaluation metrics}
To evaluate the performance of the methods, different evaluation metrics can be used. For this paper Mean Squared Error (MSE) is used as a loss function in the model. Training the model with MSE heavily punishes outliers, as the difference between prediction and real value is squared as shown in equation \ref{eq:MSE}, hence providing better results for anomaly detection.

\begin{equation}
    MSE = \frac{1}{n}\sum_{i=1}^{n}(x_i - \hat{x}_i)^{2}
    \label{eq:MSE}
\end{equation}

Where n is the number of data points predicted, $x_i$ is the real values measured, and the predicted values $\hat{x}_i$. Meanwhile in order to evaluate the outcome of the models, precision, recall and F1 scores are used. Precision score denotes the total percentage of true positives within everything detected as positive. Hence, anomalies detected by a model with high precision has more probability to be actual anomalies. Meanwhile, recall score denotes the number of true positives divided by the sum of true positives and false negatives. Thus, a model with a high recall score is better suited for finding positive cases, even though some negative cases can be categorized as positive (In the context of this research as noise). Due to individual limitations of precision and recall scores, the final evaluation on model performance was done by utilizing F1 score, which combines both scores and can be summed as the mean of precision and recall scores. The equations for the mentioned metrics are as follows:

\begin{equation}
    Precision = \frac{N_{True Positives}}{N_{True Positives} + N_{False Positives}}
    \label{eq:precision}
\end{equation}

\begin{equation}
    Recall = \frac{N_{True Positives}}{N_{True Positives} + N_{False Negatives}}
    \label{eq:recall}
\end{equation}

\begin{equation}
    F1 = 2 * \frac{Precision * Recall}{Precision + Recall}
    \label{eq:recall}
\end{equation}

\section{Methodology}
\begin{figure}[t]
    \centering
    \includegraphics[width=\linewidth]{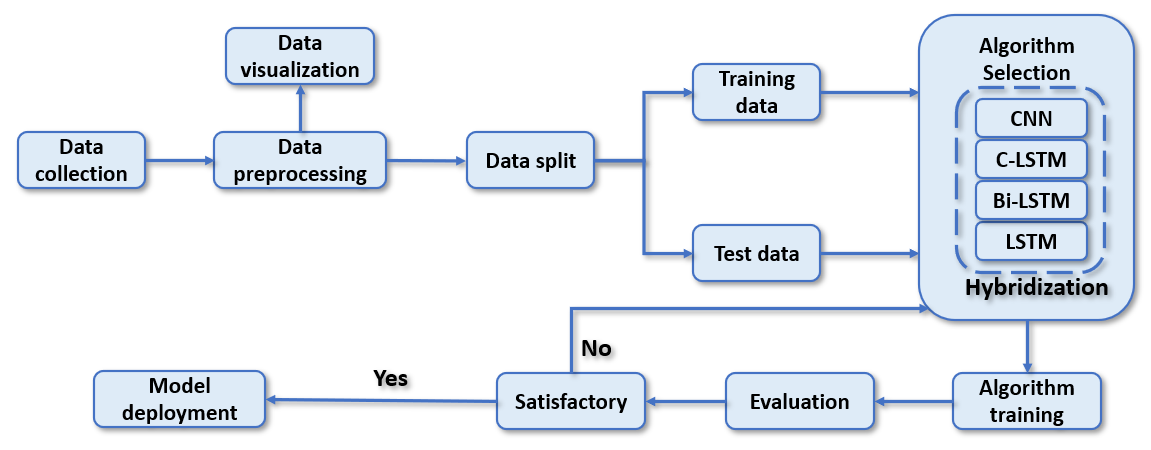}
    \caption{Methodology flowchart.}
    \label{fig:flowchart}
\end{figure}
As can be seen from Figure \ref{fig:flowchart}, the methodology of the performed research begins by data collection and preprocessing, which incorperates noise filtration, data normalization due to different scales within the data and removing of missing values if they exist. The utilized data in this research was on high quality hence no missing values have been observed.

After the initial data visualization for data preprocessing verification, the prosessed data was used for training multiple models after performing the training (80\%) and testing (20\%) data splits. The results were then evaluated based on parameters such as precision, recall and F1 scores.

\subsection{Cyber Security}
The cyber-physical interaction of power systems is a promising new technology for the future. However, the increased interaction between the cyber layer and the physical layer of the power system also presents unknown risks and challenges. Also, the amount of data generated by new technologies such as PMUs is growing rapidly. 

The increased interaction also means that the system becomes more vulnerable to cyber attacks. Hence, it is essential to detect such attacks to prevent them from causing any harm to the system. Cyber attacks that target the grid can be divided into two main categories; Cyber attacks that target the cyber layer, such as viruses, malware, and distributed denial of service (DDoS) attacks. The other category of attacks targets the physical layer, such as saboteurs and terrorists, who can physically attack parts of the grid. 

Another threat becoming increasingly common is FDIs, a type of cyber attack where false data is injected into the system. The false data can be injected into the system in various ways, for example, by tampering with the sensors measuring the grid parameters. 

Another tampering method is adversarial ML attacks, a type of cyber attack where the attackers use ML and AI algorithms to inject malicious data into the system to perturb the systems control algorithms. Adversarial machine learning attacks can be divided into two main categories: 1) Untargeted, where the attacker's goal is to distort the systems control algorithms in any way possible, and 2) targeted, where the attacker has a specific goal in mind when launching the attack. 

\subsection{Data Collection and Prepossessing }

The PMU data considered is obtained from the Texas Synchrophasor Network by the University of Texas and is available in \cite{Allen2014PMUEngineers}. The data consists of 30 Hz PMU measurements from six different location over the span of one hour and include the measurements of voltage magnitude, angle and frequency for each location. 

The data quality was observed to be high, meaning little data preprocessing is needed, as no missing values exist. Additionally, noise filtration was done through a median filter explained in \cite{Brown2016CharacterizingData} where the order of the filter determines how much of an impact it has, as a higher order includes more data points of which to find the median.

\begin{figure}[h!]
    \centering
    \includegraphics[scale=0.6]{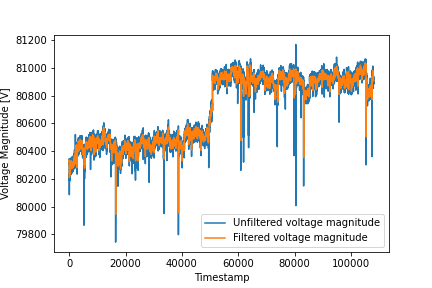}
    \caption{Filtered and unfiltered voltage measurement data}
    \label{fig:Filtered and unfiltered}
\end{figure}

The noise filtration allows the models to predict more accurately, as the variations are not as volatile. The effects of noise filtering on the data can be observed in Figure \ref{fig:Filtered and unfiltered}. The unfiltered phase voltage has a lot of minor spikes throughout, which may not be due to load variations on the grid. Meanwhile, the filtered phase voltage shows a much cleaner result where load changes are more visible. A drawback in this step can be noted as that some of the major spikes disappear, since they are too brief to have an impact in the filtering process. 

\begin{figure}[h!]
    \centering
    \includegraphics[scale=0.55]{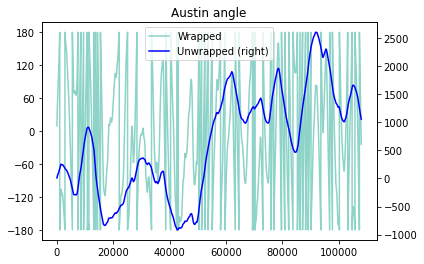}
    \caption{Wrapped and Unwrapped voltage angle measurements.}
    \label{fig:angle}
\end{figure}

Additionally, the voltage angle values within the dataset were unwrapped, since unwrapped angles are needed for the machine learning algorithms to "learn" the real difference in angle from measurement to measurement, as the jump between +180 to -180 degrees is solely conventional and has no true meaning for the behaviour of the power grid. Thus, the need for unwrapping before creating train and test sets for the model development part. The wrapped and unwrapped voltage angle values are illustrated in Figure \ref{fig:angle}.

As a next step, the dataset was normalized in order to eliminate different ranges within different features, which was done via scikit-learn python package. 

\subsection{Model Development }

Four models were developed for FID anomaly detection within this study were; CNN, LSTM, Bi-LSTM and C-LSTM where the latter two are hybrid models. The input parameters to the layers of the models are optimized using the Talos optimizer python package. The best parameters for anomaly detection purposes with the proposed models are shown in Table \ref{tab:modelparams}  for the CNN, LSTM and Bi-LSTM, and C-LSTM models respectively.

\begin{table}[]
\resizebox{9.5cm}{2.8 cm}{%
\centering

\begin{tabular}{|ccc|ccc|ccc|}
\hline
\multicolumn{3}{|c|}{\textbf{{ CNN}}}                                                      & \multicolumn{3}{c|}{\textbf{ LSTM/Bi-LSTM}}                                                                                       & \multicolumn{3}{c|}{\textbf{ C-LSTM}}                                                                                        \\ \hline
\multicolumn{1}{|c|}{\multirow{3}{*}{ Conv1D}} & \multicolumn{1}{c|}{ Filters}     & { 32}   & \multicolumn{1}{c|}{\multirow{3}{*}{\shortstack{ LSTM \\ Bi-LSTM}}} & \multicolumn{1}{c|}{\multirow{3}{*}{ \shortstack{Hidden\\ Units}}}  & \multirow{3}{*}{ 64}   & \multicolumn{1}{c|}{\multirow{3}{*}{ Conv1D}}  & \multicolumn{1}{c|}{ Filters}                        & { 64}                    \\ \cline{2-3} \cline{8-9} 
\multicolumn{1}{|c|}{}                        & \multicolumn{1}{c|}{ \shortstack{Kernel\\ size}} & { 3 }   & \multicolumn{1}{c|}{}                              & \multicolumn{1}{c|}{}                               &                       & \multicolumn{1}{c|}{}                         & \multicolumn{1}{c|}{ \shortstack{Kernel\\ Size}}                    & { 3}                     \\ \cline{2-3} \cline{8-9} 
\multicolumn{1}{|c|}{}                        & \multicolumn{1}{c|}{ Activation}  & { relu} & \multicolumn{1}{c|}{}                              & \multicolumn{1}{c|}{}                               &                       & \multicolumn{1}{c|}{}                         & \multicolumn{1}{c|}{ Activation}                     & { relu}                  \\ \hline
\multicolumn{1}{|c|}{\multirow{3}{*}{ Conv1D}} & \multicolumn{1}{c|}{ Filters}     & { 16}   & \multicolumn{1}{c|}{\multirow{3}{*}{ Dropout}}      & \multicolumn{1}{c|}{\multirow{3}{*}{ \shortstack{Dropout\\ Value}}} & \multirow{3}{*}{ 0.12} & \multicolumn{1}{c|}{\multirow{3}{*}{ LSTM}}    & \multicolumn{1}{c|}{\multirow{3}{*}{ \shortstack{Hidden\\ Units}}}  & \multirow{3}{*}{ 64}   \\ \cline{2-3}
\multicolumn{1}{|c|}{}                        & \multicolumn{1}{c|}{ \shortstack{Kernel\\ size}} & { 3}    & \multicolumn{1}{c|}{}                              & \multicolumn{1}{c|}{}                               &                       & \multicolumn{1}{c|}{}                         & \multicolumn{1}{c|}{}                               &                       \\ \cline{2-3}
\multicolumn{1}{|c|}{}                        & \multicolumn{1}{c|}{ Activation}  & { relu} & \multicolumn{1}{c|}{}                              & \multicolumn{1}{c|}{}                               &                       & \multicolumn{1}{c|}{}                         & \multicolumn{1}{c|}{}                               &                       \\ \hline
\multicolumn{1}{|c|}{\multirow{3}{*}{ Conv1D}} & \multicolumn{1}{c|}{ Filters}     & { 64}   & \multicolumn{1}{c|}{\multirow{3}{*}{ LSTM}}         & \multicolumn{1}{c|}{\multirow{3}{*}{ \shortstack{Hidden\\ Units}}}  & \multirow{3}{*}{ 32}   & \multicolumn{1}{c|}{\multirow{3}{*}{ Dropout}} & \multicolumn{1}{c|}{\multirow{3}{*}{ \shortstack{Dropout\\ Value}}} & \multirow{3}{*}{ 0.12} \\ \cline{2-3}
\multicolumn{1}{|c|}{}                        & \multicolumn{1}{c|}{ \shortstack{Kernel\\ size}} & { 3}    & \multicolumn{1}{c|}{}                              & \multicolumn{1}{c|}{}                               &                       & \multicolumn{1}{c|}{}                         & \multicolumn{1}{c|}{}                               &                       \\ \cline{2-3}
\multicolumn{1}{|c|}{}                        & \multicolumn{1}{c|}{ Activation}  & { relu} & \multicolumn{1}{c|}{}                              & \multicolumn{1}{c|}{}                               &                       & \multicolumn{1}{c|}{}                         & \multicolumn{1}{c|}{}                               &                       \\ \hline
\multicolumn{1}{|c|}{ \shortstack{Max\\Pooling}}              & \multicolumn{1}{c|}{ \shortstack{Pool\\ size}}   & { 2}    & \multicolumn{1}{c|}{ Dense}                         & \multicolumn{1}{c|}{ Units}                          & { 1}                     & \multicolumn{1}{c|}{ LSTM}                     & \multicolumn{1}{c|}{ \shortstack{Hidden\\ Units}}                   & { 32}                    \\ \hline
\multicolumn{1}{|c|}{\multirow{2}{*}{ Dense}}  & \multicolumn{1}{c|}{ Units}       & { 50}   & \multicolumn{1}{c|}{\multirow{2}{*}{}}             & \multicolumn{1}{c|}{\multirow{2}{*}{}}              & \multirow{2}{*}{}     & \multicolumn{1}{c|}{\multirow{2}{*}{ Dropout}} & \multicolumn{1}{c|}{\multirow{2}{*}{ \shortstack{Dropout\\ value}}} & \multirow{2}{*}{ 0.12} \\ \cline{2-3}
\multicolumn{1}{|c|}{}                        & \multicolumn{1}{c|}{ Activation}  & { relu} & \multicolumn{1}{c|}{}                              & \multicolumn{1}{c|}{}                               &                       & \multicolumn{1}{c|}{}                         & \multicolumn{1}{c|}{}                               &                       \\ \hline
\multicolumn{1}{|c|}{ Dense}                   & \multicolumn{1}{c|}{ Units}       & { 1}    & \multicolumn{1}{c|}{}                              & \multicolumn{1}{c|}{}                               &                       & \multicolumn{1}{c|}{ Dense}                    & \multicolumn{1}{c|}{ Units}                          & { 1}                     \\ \hline
\multicolumn{1}{|c|}{ Flatten}                 & \multicolumn{1}{c|}{ -}           & { -}    & \multicolumn{1}{c|}{}                              & \multicolumn{1}{c|}{}                               &                       & \multicolumn{1}{c|}{}                         & \multicolumn{1}{c|}{}                               &                       \\ \hline
\end{tabular}%
}
\\
\caption{\quad Model parameters after hyperparameter optimization.}
\label{tab:modelparams}
\end{table}

All the models, except for CNN, largely follow a similar structure with two main LSTM-based layers, each followed by a dropout layer to reduce effects of overfitting, before a dense layer is added to reduce the  dimensionality of the output. Meanwhile, the C-LSTM model has an additional convolutional layer as its first layer to integrate its properties and hybridize the model.

\subsection{Model calibration}

To ensure that the proposed models are reliable and give accurate predictions, they were calibrated by the statistical methods used in \cite{Allen2014PMUEngineers} where the authors of the study propose a script using statistical methods to find real events in the PMU data used in this paper. Deployment of the methods proved that the same anomalies were detected using the above proposed models and the statistical methods of the authors.

\subsection{Data injection}

To mimic a sensor malfunction or a white noise attack on the system, a Gaussian noise is added to a small portion of the data. This noise is added at a random location and lasts for a duration between one and two seconds at every given time.

\section{Results}

To evaluate the efficacy of the models, using the appropriate criteria and metrics are important. For anomaly detected with injected data, accuracy is not a good indicator, as the anomalous data points are in vast minority to the total amount of data. Therefore, the metrics used for comparison are precision, recall and F1 score.

\begin{table}[b]
    \centering
    \begin{tabular}{|c|c|c|c|c|} \hline
       \multirow{2}{*}{\textbf{Model}} & \textbf{Noise} & \multirow{2}{*}{\textbf{Recall (\%)}} & \multirow{2}{*}{\textbf{Precision (\%)}} & \multirow{2}{*}{\textbf{F1-Score(\%)}} \\
                              & \textbf{filtration} & & & \\\hline
        \multirow{2}{*}{CNN} & Yes & 94.67 & 87.65 & 91.03 \\ \cline{2-5}
                                 & No & 86.67 & 94.20 & 90.28\\ \hline
        \multirow{2}{*}{LSTM} & Yes & 90.38 & 94.0 & 92.16 \\ \cline{2-5}
                                 & No & 91.23 & 69.33 & 78.79\\ \hline
        \multirow{2}{*}{Bi-LSTM} & Yes & 94.05 & \textbf{98.75} & 96.34 \\ \cline{2-5}
                                 & No & 82.0 & 91.11 & 86.32\\ \hline
        \multirow{2}{*}{C-LSTM} & Yes & \textbf{97.50} & 96.30 & \textbf{96.89} \\ \cline{2-5}
                                 & No & 94.12 & 96.0 & 95.05\\ \hline 
    \end{tabular}
    \caption{Performance results for the different models}
    \label{tab:results}
\end{table}

\begin{figure}[t]
    \centering
    \includegraphics[scale=0.55]{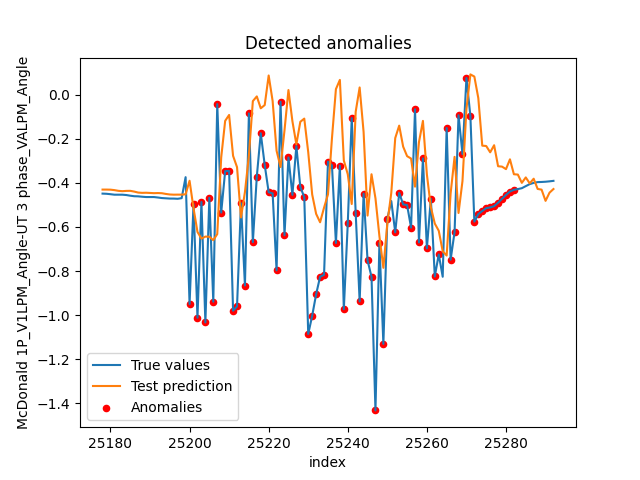}
    \caption{The CNN model without noise filtration.}
    \label{fig:zoom_both_cnn_noise}
\end{figure}

The results are presented in Table \ref{tab:results}. As can be seen, the results for the different models vary greatly in their performance and between metrics. The CNN model with noise filtration seems to label too many data points as anomalies and therefore suffers in both precision and F1-score. However, for the same reason, there are few false negatives and provides a recall of 94.20\%. Meanwhile, the CNN model without noise filtration catches too few anomalies which is due to the fact that more noise in the training data will result in the model setting a higher anomaly threshold. From Figure \ref{fig:zoom_both_cnn_noise}, it is also clear that some false positives occur after the injected noise since the CNN model requires more time to recover from the injected noise.

\begin{figure}
    \centering
    \includegraphics[scale=0.55]{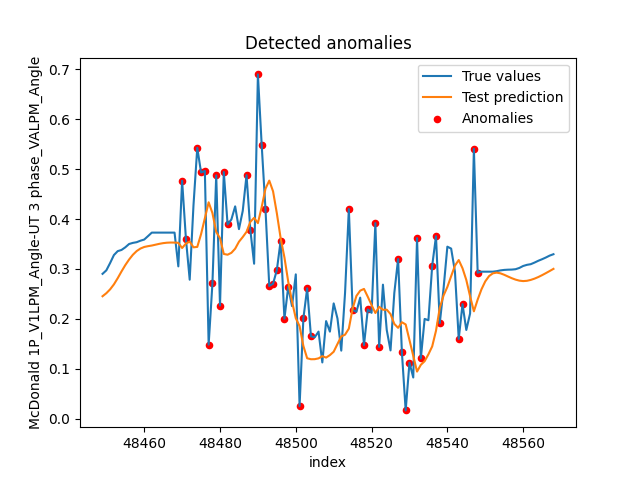}
    \caption{LSTM - Detected anomalies with pred with noise filter}
    \label{fig:LSTManomswithNoise}
\end{figure}

\begin{figure}
    \centering
    \includegraphics[scale=0.55]{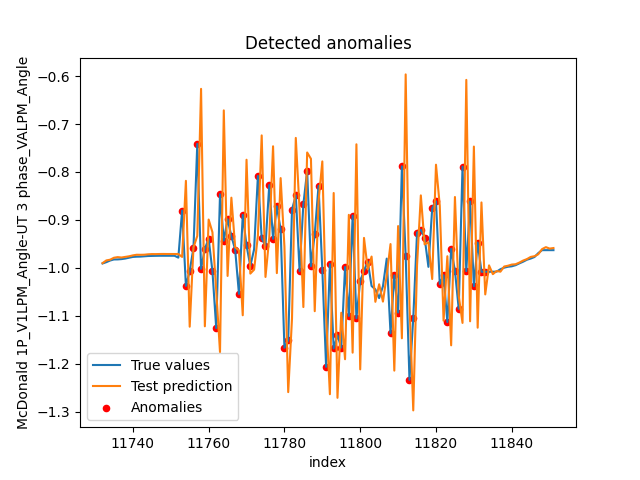}
    \caption{C-LSTM - Detected anomalies with noise filtration.}
    \label{fig:CLSTManomswithNoise}
\end{figure}

\begin{figure}
    \centering
    \includegraphics[scale=0.55]{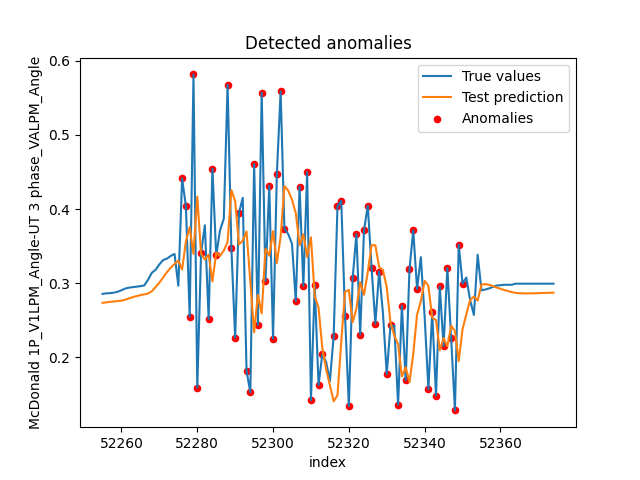}
    \caption{Bi-LSTM - Detected anomalies with noise filtration.}
    \label{fig:biLSTManomswithNoise}
\end{figure}

For the LSTM-based models, noise filtration is essential for their ability for high performance anomaly detection. From Table \ref{tab:results} it can be observed that the LSTM model outperforms the CNN model with noise filtration, however does comparatively worse without noise filtration. The LSTM model seems to have the highest benefit of noise filtration having a jump in F1-score from 78.79\% up to 92.16\%. In Figure \ref{fig:LSTManomswithNoise} and Figure \ref{fig:biLSTManomswithNoise} its clear that both pure and Bi-LSTM have overall a bad fit compared to the ground truth data. However, they still perform with F-1 scores above 90\%. Meanwhile, hybritizing the LSTM model by coupling it with CNN offers adventages of both algorithms. In Figure \ref{fig:CLSTManomswithNoise} the anomaly detection results of the hybrid algorithm can be seen. Even though, the predictions done by the model have clear overshoots with respect to the ground truth data. The model outcome is the best between all the assessed models with and without noise filtration with an F-1 score of 96.89\%.

\section{Discussion and Conclusion}

As can be seen from the results, the labelling of anomalies is a challenge. The anomaly detection relies heavily on the computed threshold. Thus, finding the optimal threshold considering the nature of the optimal power system utilization and injected data properties is essential. One way to achieve this was found to be utilization of MSE when the model predicts the training data. As can be observed, this method was found out to give satisfactory results with F-1 scores ranging from 78.79\% to 96.89\%. However, for future work, a dynamic anomaly detection threshold can be implemented to observe its effects, instead of a single global threshold for whole dataset. Additionally, models can be trained for detecting other anomaly types such as sensor drifts, spikes and surges. 

In this research, various AI models were trained and assessed in order to detect anomalies in PMU datasets, which can arise after a FDI cyber-attack. The results of the study might provide help towards various Transmission System Operators (TSOs) in order to better secure the grid and provide stable operation.

\bibliographystyle{IEEEtran}
\bibliography{references}
\end{document}